\documentclass{article}
\usepackage{ijcai24}

\usepackage{times}
\usepackage{soul}
\usepackage{url}
\usepackage[hidelinks]{hyperref}
\usepackage[utf8]{inputenc}
\usepackage[small]{caption}
\usepackage{graphicx}
\usepackage{amsmath}
\usepackage{amsthm}
\usepackage{booktabs}
\usepackage{algorithm}
\usepackage{algorithmic}

\title{Challenging the Human-in-the-loop in Algorithmic Decision-making}

\author{
Sebastian Tschiatschek$^{1}$
\and
Eugenia Stamboliev$^{2}$\and
Timothée Schmude$^{1}$\and\\
Mark Coeckelbergh$^{2}$\And
Laura Koesten$^{1}$\\
\affiliations
$^1$University of Vienna, Faculty of Computer Science\\
$^2$University of Vienna, Department of Philosophy\\
}

\usepackage[nohyperlinks, nolist]{acronym}
\usepackage{verbatim}
\usepackage{amsmath,amssymb,amsthm}
\usepackage{booktabs}
\usepackage{fontawesome5}

\usepackage{tikz}
\usetikzlibrary{fit,calc,backgrounds,positioning,patterns}

\usepackage{caption}
\usepackage{subcaption}
\usepackage[inline]{enumitem}

\usepackage{ifthen}
\newboolean{include-notes}
\setboolean{include-notes}{true}
\newcommand{\nb}[3]{\ifthenelse{\boolean{include-notes}}{{\colorbox{#2}{\bfseries\sffamily\scriptsize\textcolor{white}{#1}}}{\ \textcolor{#2}{\sf\small\textit{#3}}}}{}}

\newcommand{\alg}{\ensuremath{\mathcal{A}}}
\newcommand{\algs}{\ensuremath{\mathbb{A}}}

\newcommand{\statistics}{\ensuremath{\rho}}

\tikzstyle{var}=[draw=blue!50!black,fill=blue!20!white,circle,minimum width=10mm]
\tikzstyle{varH}=[draw=blue!50!black,fill=white!20!white,circle,minimum width=10mm]
\tikzstyle{arrow}=[->,>=latex,line width=0.3mm]

\definecolor{pdm}{HTML}{4FB4CE}
\definecolor{sdm}{HTML}{E86654}

\tikzstyle{exp}=[draw opacity=0.5, dotted, line width=2pt]
\colorlet{exp1color}{red}
\tikzstyle{exp1}=[exp, draw=exp1color!80!white]
\colorlet{exp2color}{green}
\tikzstyle{exp2}=[exp, draw=exp2color!80!white]
\colorlet{exp3color}{blue}
\tikzstyle{exp3}=[exp, draw=exp3color!80!white]
\colorlet{exp4color}{orange}
\tikzstyle{exp4}=[exp, draw=exp4color!80!white]

\tikzset{fit margins/.style={/tikz/afit/.cd,#1,
    /tikz/.cd,
    inner xsep=\pgfkeysvalueof{/tikz/afit/left}+\pgfkeysvalueof{/tikz/afit/right},
    inner ysep=\pgfkeysvalueof{/tikz/afit/top}+\pgfkeysvalueof{/tikz/afit/bottom},
    xshift=-\pgfkeysvalueof{/tikz/afit/left}+\pgfkeysvalueof{/tikz/afit/right},
    yshift=-\pgfkeysvalueof{/tikz/afit/bottom}+\pgfkeysvalueof{/tikz/afit/top}},
    afit/.cd,left/.initial=2pt,right/.initial=2pt,bottom/.initial=2pt,top/.initial=2pt}

\newcommand{\w}{\ensuremath{\mathbf{w}}}
\newcommand{\x}{\ensuremath{\mathbf{x}}}

\newtheorem{observation}{Observation}

\begin{document}

\begin{acronym}
  \acro{adm}[ADM]{algorithmic decision-making}
  \acro{hil}[HIL]{human-in-the-loop}
  \acroplural{hil}[HILs]{humans-in-the-loop}
  \acro{pdm}[PDM]{practical decision-maker}
  \acro{sdm}[SDM]{strategic decision-maker}
  \acro{xai}[XAI]{explainable AI}
\end{acronym}

\maketitle

\begin{abstract}
  We discuss the role of humans in \ac{adm} for socially relevant problems from a technical and philosophical perspective.
  In particular, we illustrate tensions arising from diverse expectations, values, and constraints by and on the humans involved.
  To this end, we assume that a \ac{sdm} introduces \ac{adm} to optimize strategic and societal goals while the algorithms' recommended actions are overseen by a \ac{pdm} -- a specific \acl{hil} -- who makes the final decisions.
  While the \ac{pdm} is typically assumed to be a corrective, it can counteract the realization of the \ac{sdm}'s desired goals and societal values not least because of a misalignment of these values and unmet information needs of the \ac{pdm}.
  This has significant implications for the distribution of power between the stakeholders in \ac{adm}, their constraints, and information needs.
  In particular, we emphasize the overseeing \ac{pdm}'s role as a potential political and ethical decision maker, who acts expected to balance strategic, value-driven objectives and on-the-ground individual decisions and constraints.
  We demonstrate empirically, on a machine learning benchmark dataset, the significant impact an overseeing \ac{pdm}'s decisions can have even if the \ac{pdm} is constrained to performing only a limited amount of actions differing from the algorithms' recommendations.
  To ensure that the \ac{sdm}'s intended values are realized, the \ac{pdm} needs to be provided with appropriate information conveyed through tailored explanations and its role must be characterized clearly. 
  Our findings emphasize the need for an in-depth discussion of the role and power of the \ac{pdm} and challenge the often-taken view that just including a \acl{hil} in ADM ensures the `correct' and `ethical' functioning of the system.

\end{abstract}

\section{Introduction}

The application of \acf{adm} in public services and businesses is increasing rapidly~\cite{european_commission_laying_2021}.
As a result, algorithms have a significant influence on us as individuals and society~\cite{veale_2018_fairness}. 
Concrete example applications include the usage of \ac{adm} for the provision of support measures in the labor market~\cite{scott_algorithmic_2022}, college admission, credit application decisions \cite{wachter2017right}, refugee settlement~\cite{Bansak2018}, recidivism prediction~\cite{dressel2018accuracy,chouldechova2017}, or child welfare services~\cite{brown_toward_2019}.

While potential benefits of \ac{adm} have been praised---scalability, reproducibility, and fairness---, real-world applications have repeatedly been criticized for being non-transparent, unreliable, biased, and consequently harmful~\cite{barocas2016big,brown_toward_2019,chouldechova2017,definelicht2020,bell2022itsjust,woodruff_qualitative_2018}.
One popular way to counteract these shortcomings is to ensure human oversight~\cite{hermstruwer2022fair} of \ac{adm} as a safeguard against the pitfalls of automation~\cite{high_level_expert_group_on_ai_eu,eubanks2018automating}. 
In this work, we discuss the use of \ac{adm} with societal impact exemplified in the case of employment service algorithms in which individual organizations and regulators increasingly aim to take precautions ~\cite{allhutter_bericht_ams-algorithmus_2020,seidelin2022auditing,scott_algorithmic_2022}.

The human overseers are commonly viewed as controlling the \ac{adm} to guarantee accountability but also to prevent unpredictable and unpredicted consequences~\cite{dodge_explaining_2019,starke2022fairness_perceptions,rahwan2018society,dg2019understanding}.
Despite other aspects, the concrete impact of \ac{adm} on individuals and society, therefore, depends on human actors along the line, the \emph{\aclp{hil}} (\acsp{hil}).
\acused{hil}

Defining \acp{hil}, the machine learning literature commonly focuses on the integration of human knowledge and experience directly in a predictive model.
In the context of ADM this could concern data preprocessing or annotation, or different feedback loops within a model~\cite{DBLP:journals/fgcs/WuXSZM022}.
However, especially in \ac{adm} for public services, there are also \acp{hil} outside the direct algorithmic process who influence the outcome. In this work, we put the focus on these other \acp{hil}:
\begin{enumerate*}[label=(\roman*), font=\itshape]
  \item the person who makes a final decision informed by the system’s predictions \cite{bell2022itsjust}---the \emph{\ac{pdm}} for the purpose of this work; and
  \item the people who have strategic oversight over the development and deployment of an \ac{adm} system, such as politicians, regulators, or other actors in society who determine longer-term goals in the context of public services---the \emph{\acf{sdm}} in this work.
\end{enumerate*}
These two \acp{hil} within an \ac{adm} process are illustrated in Figure~\ref{fig:framework-human-readble}.

While the importance of human actors in \ac{adm} for public services has been recognized, we argue that this consequently requires rethinking the roles and power of the different \acp{hil} involved.
This includes in-depth considerations of the work practices where the models’ predictions are meant to be used~\cite{DBLP:conf/ACMdis/DhanorkarWQXP021}, decision-making powers, value alignment, and information needs of these key stakeholders~\cite{langer_what_2021}.
Furthermore, complex predictive algorithms are not always intelligible to
humans, posing both practical and ethical concerns ~\cite{rudin2019stop}. 
Different stakeholders have different information needs~\cite{shulner-tal_enhancing_2022,dodge_explaining_2019}, depending on their type of involvement in the \ac{adm} process, ranging from development and deployment to regulation of an \ac{adm} system~\cite{arrietta2020,langer_what_2021}.
While the need for technical explanations has been a key topic in the recent \ac{xai} literature~\cite{speith_review_2022,arrieta2020explainable}, the need to also understand social, societal, and ethical implications of an \ac{adm} process by all \acp{hil} has received significantly less attention.
We argue that the challenge of weighing up individual decisions against strategic goals needs increased attention, considering the crucial role of both the \ac{sdm} and the \ac{pdm} in \ac{adm} for public services.

The \acp{hil} must decide smartly, responsibly, and ethically, not only over the algorithm but over the \ac{adm}'s consequences and that of the applied treatment.
Use cases surface a variety of practical tensions related to value alignment, constraints of power and workload, and only relatively vague concepts of the skills and knowledge required of the \acp{hil}.
We aim to highlight, that the \ac{pdm}'s decision can in fact change the desired behavior of \ac{adm}.
This is partly in tension with recent research which mainly identifies \acp{hil} as key for ensuring safe usage of ADM and observes increased trust in systems with \acp{hil}~\cite{Hidalgo2021}.

In this work, we discuss the challenges of the \ac{pdm} as a technical, political, and ethical decision-maker, who acts in a field of tension between strategic objectives and individual decisions and constraints.  
Our contributions are as follows: 

\begin{figure}[tbp]
    \centering
    \includegraphics[]{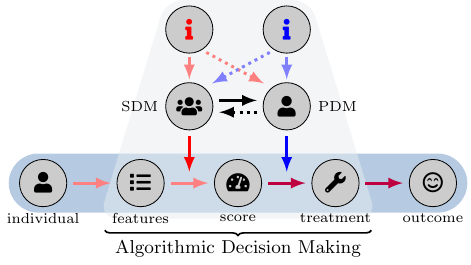}
    \vspace{-2mm}
    
    \caption{Considered \ac{adm} process.
    The \ac{sdm} makes long-term decisions regarding the process, e.g., which societal values it should realize, while the \ac{pdm} oversees decisions on individuals.
    The \ac{adm} makes recommendations for treatments of individuals based on an individual's score predicted from an individual's features.
    The actual applied treatment is selected by the \ac{pdm} and results in an outcome for the individual.
    To realize the \ac{sdm}'s intended values, the \ac{sdm} and the \ac{pdm} need to account for each other and act accordingly while both might base their decisions on different information available to them.
    }
    
    \label{fig:framework-human-readble}
    \vspace{-4mm}
\end{figure}

\begin{enumerate}
    \item We conceptualize \ac{adm} for socially relevant problems with respect to the goals and roles of two key stakeholders to foster a philosophical and theoretical investigation and shed light on neglected topics in this research area.

    \item We highlight the power of the \ac{pdm}, the tension between them and the \ac{sdm}, and differences in their information needs.
    \item Experimentally we demonstrate the impact of misalignment between the \ac{sdm} and the \ac{pdm} regarding the realization of the societal goals and the challenges the \ac{pdm} faces regarding supporting the strategic goals.
    \item We connect our conceptualization to a philosophical discourse about what needs to be explained and how power and strategic/political control manifest.

\end{enumerate}

\section{Related Work}
\label{sec:related}

Due to space constraints, we only present a selection of related work here---see the appendix for further related work.

The setting considered in this paper is closely related to the \emph{principal-agent problem}~\cite{eisenhardt1989agency}, which characterises conflicts of interests and priorities that can arise if an agent (in our case the \ac{pdm}) acts on behalf of another identity (in our case the \ac{sdm}).
As such, considerations of issues of information and power asymmetries are central to the problem.
Our paper can be considered as studying the pecularities of the principal-agent problem in settings involving \ac{adm}, putting related work in context, and proposing approaches for cases in which agents are supported by algorithms.

\paragraph{Humans in the loop.}
A \emph{human-in-the-loop} setting is commonly defined as an automated process that requires human interaction, meaning that human knowledge and experience are integrated into, for instance, an ML model~\cite{DBLP:journals/fgcs/WuXSZM022}.
Some authors also differentiate the \emph{human-on-the-loop}/\emph{human oversight} process in which the human perform a less central monitoring or supervisory role~\cite{fischer2021loop}.
Oversight is commonly discussed in the context of trust in \ac{adm} and AI and which is increasingly required by policy-makers and regulators~\cite{koulu2020human,high_level_expert_group_on_ai_eu,europeancommission_2021_artifical}. 

The \ac{hil} is commonly viewed as ensuring accountability and preventing undesirable consequences~\cite{dodge_explaining_2019,starke2022fairness_perceptions,high_level_expert_group_on_ai_eu,rahwan2018society}. 
Algorithmic accountability has been established to describe good practice for \ac{adm} use in public services~\cite{brown_toward_2019}, often attributed to the existence of a \ac{hil}.
However, the traditional \ac{hil} %
does not necessarily fit this role.
Usually, the focus is on increasing the performance of the algorithm or resolving ambiguities~\cite{DBLP:journals/fgcs/WuXSZM022}.
A large fraction of research on \ac{adm} does not even specify the precise role of the \ac{hil}, e.g., whether a \ac{hil} has the power to influence the \ac{adm} outcome and how this would affect the environment. 
However, questions regarding the distribution of power have been raised~\cite{kasy2021fairness}.

Another important related work is~\cite{green2019principles}. It differs from our paper in the basic setting, i.e., it assumes a setting in which there are correct ethical decisions and investigates how people interact with an algorithm to realize them. In contrast, in our setting, the values to realize are not globally normative but based on the decisions of the \acp{sdm} and hence our focus is more on how the alignment of \acp{sdm} and \acp{pdm} can be achieved and understood.

\paragraph{Information needs of humans in the loop.}
Different \acp{hil} need different types of knowledge about the \ac{adm} process depending on their role~\cite{langer_what_2021}. To date, most \ac{xai} literature focuses on the technicalities of automated processes~\cite{speith_review_2022,abdul2018trends}. 
While recent work and guidelines demonstrate increasing awareness of the potential pitfalls and dangers of \ac{adm}~\cite{brown_toward_2019,definelicht2020}, many proposed solutions are designed with data scientists or developers in mind, closely linking explanations with the model process~\cite{DBLP:conf/chi/LiaoGM20}. 
This common, algorithmic-centric view of information needs focuses on information about input data and potential biases, performance, feature weights, and other aspects of \ac{adm}~\cite{DBLP:conf/chi/LiaoGM20}.
While relevant, a need for other diverse explanations has been recognized. Within the technical realm this includes considering `what' is explained; for instance, differentiating between \emph{local} (a predictive output is explained) and \emph{global}
(concerning the broader overall reasoning of the model) explanations
\cite{DBLP:conf/ACMdis/DhanorkarWQXP021}.
What is more, this also includes considering `who' explanations are tailored to, as different stakeholders are involved in \ac{adm}~\cite{langer_what_2021}.
Some effects of explanations on AI-human teams have been considered in~\cite{bansal2021does}, although with a different focus.
The authors observed that explanations increased the chance that humans accept AI recommendations independent of their correctness, emphasizing the need to reflect upon work practices and tensions arising between AI and \acp{hil}.
This work goes further in illustrating the needs of the \acp{hil}, widening the scope of the required information, skills, and explanations.

\section{Conceptualization}
\label{sec:conceptualization}

We aim to understand the distribution of power between the \ac{sdm} and \ac{pdm} in \ac{adm} and their diverse information needs. %
To this end, we first describe a real-world motivating example followed by its conceptualization.

\subsection{A Motivating Example}%
\label{sec:motivating-example}

As a motivating example we consider employment service algorithms according to \cite{allhutter_bericht_ams-algorithmus_2020,seidelin2022auditing,scott_algorithmic_2022}. 
In this example, the \ac{adm} systems are used to decide on actions concerning job seekers, e.g., providing funding for attending courses that improve a job seekers' qualifications, based on predictions of the job-seekers' employability.
The prediction is based on various features of a job-seeker, e.g., their age, gender, and care duties. 
Based on the prediction and overall statistics of the population, a treatment (action) is recommended by the employed algorithm.
This treatment recommendation is reviewed by the \ac{pdm}, who is a domain expert for the labor market but not for \ac{adm}.
The \ac{pdm} makes the final treatment decision.
Importantly, the \ac{pdm} is not a single person but corresponds to many persons across the organization, e.g., each \ac{pdm} is active in a specific geographic region.%

The overall decision for implementing the \ac{adm} system as described above has been made by the \ac{sdm} with goals of societal relevance, e.g., decreasing unemployment rates.

While the described running example considers a specific use case, the same structure can be observed in other domains, e.g., recidivism prediction and refugee resettlement.
The sketched process is illustrated in Figure~\ref{fig:framework-human-readble}.
In this figure, we particularly highlight that the decisions of the \ac{sdm} and the \ac{pdm} are based on general information as well as an exchange of information between them. %

\subsection{Framework}
\label{sec:framework}

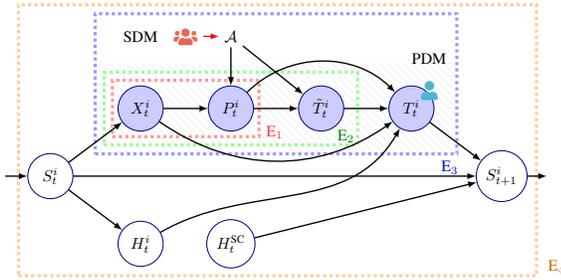
\begin{figure}
    \centering
        \scalebox{0.60}{\begin{tikzpicture}
  \node[varH] (S_t) {$S^i_t$};
  \node[var, right of=S_t, xshift=10mm, yshift=15mm] (X_t) {$X^i_t$};
  \node[varH, right of=S_t, xshift=10mm, yshift=-15mm] (H_t) {$H^i_t$};
  \node[varH, right of=H_t, xshift=10mm] (HSC_t) {$H_t^\textnormal{SC}$};
  \node[var, right of=X_t, xshift=10mm] (P_t) {$P^i_t$};
  \node[var, right of=P_t, xshift=10mm] (Ttilde_t) {$\tilde{T}^i_t$};
  \node[var, right of=Ttilde_t, xshift=10mm] (T_t) {$T_t^i$};
  \node[xshift=4mm,yshift=4mm] at (T_t) (p_2) {\Large\color{pdm}\faUser};
  \node[above of=p_2,anchor=east, xshift=5mm, yshift=-3mm] () {\ac{pdm}};
  \node[varH, right of=T_t, xshift=10mm, yshift=-15mm] (S_tp) {$S^i_{t+1}$};
  
  \draw[arrow] ($(S_t) + (-10mm, 0mm)$) -- (S_t);
  \draw[arrow] (S_t) -- (X_t);
  \draw[arrow] (S_t) -- (H_t);
  \draw[arrow] (X_t) -- (P_t);
  \draw[arrow] (P_t) -- (Ttilde_t);
  \draw[arrow] (Ttilde_t) -- (T_t);
  \draw[arrow] (P_t) to[out=45] (T_t);
  \draw[arrow] (X_t) to[out=-35,in=-145] (T_t);
  \draw[arrow] (T_t) -- (S_tp);
  \draw[arrow] (S_t) -- (S_tp);
  \draw[arrow] (H_t) to[out=30, in=-120] (T_t);
  \draw[arrow] (HSC_t) -- (S_tp);
  \draw[arrow] (S_tp) -- ($(S_tp) + (10mm, 0mm)$);

  \node[above of=P_t, yshift=6mm] (alg) {$\alg$};
  \node[left of=alg] (p1) {\Large\color{sdm}\faUsers};
  \node[left of=p1,anchor=east, xshift=5mm] () {\ac{sdm}};
  \draw[arrow, thick, draw=red] (p1) -- (alg);
  \draw[arrow] (alg) -- (Ttilde_t);
  \draw[arrow] (alg) -- (P_t);
  
  \begin{scope}[on background layer]
      \node[exp4, fit=(X_t)(T_t),inner sep=5mm, pattern=north west lines, pattern color=gray!20!white, draw=none] () {};
  
      \node[exp1, fit=(X_t)(P_t), label={[xshift=0mm,yshift=4mm]south east:\textcolor{exp1color!80!white}{$\textnormal{E}_1$}}] () {};
      
      \node[exp2, fit=(X_t)(Ttilde_t), inner sep=3mm, label={[xshift=-6mm,yshift=5mm]south east:\textcolor{exp2color!50!black}{$\textnormal{E}_2$}}] () {};

      \node[exp3, fit={(X_t)(T_t)($(p1)+(0mm,0mm)$)},inner sep=5mm, label={[xshift=-5mm]south east:\textcolor{exp3color!80!black}{$\textnormal{E}_3$}}] () {};
      
      \node[exp4, inner sep=0mm, fit margins={left=1mm,right=1mm,bottom=1mm,top=3.5mm}, fit={(S_t)(X_t)(H_t)(S_tp)($(p1)+(0mm,0mm)$)}, label={[xshift=1mm, yshift=5mm]south east:\textcolor{exp4color!80!black}{$\textnormal{E}_4$}}] () {};

  \end{scope}

\end{tikzpicture}}
    \vspace{-2mm}

    \caption{\ac{adm} framework with \emph{\acl{sdm}} (\scalebox{0.8}{\color{sdm}\faUsers}) and \emph{\acl{pdm}} (\scalebox{0.8}{\color{pdm}\faUser}).
    Explanations for different parts of the framework are marked by colored boxes.
    The circled nodes represent the true state of an individual $S_t^i$, its observed features $X_t^i$ and confounding factors $H_t^i$, the prediction by an algorithm $\alg$ for the individual $P_t^i$, the suggested treatment $\tilde{T}_t^i$ according to the algorithm, and the next state of the individual $S_{t+1}^i$.
    This next state also depends on hidden confounders $H_t^\textnormal{SC}$ independent of an individual, e.g., changes in laws. 
    The \ac{pdm} can alter the treatment, becoming $T_t^i$.
    Algorithm $\alg$ is selected to realize societal and ethical values. %
    }
    
    \label{fig:framework}
    \vspace{-4mm}
\end{figure}

\paragraph{Overview.}
In our conceptualization, we consider a (sequential) decision-making problem that is governed by \ac{adm}.
Concretely, we consider the following setting (cf.\ Figure~\ref{fig:framework}):
\begin{enumerate*}[label=(\alph*), font=\itshape]
    \item %
      \ac{adm} is used to make decisions about individuals.

    \item \ac{adm} is used with \acp{pdm} to increase trust and reliability.
    The \acp{pdm} make the final decisions and can take the recommendations of the algorithm into account. %
    
    \item The \ac{sdm} decides on which algorithm $\alg$ to use inside \ac{adm}, e.g., by selecting from a set of algorithms $\algs$. 
      The algorithm is selected to be best \emph{aligned with the values} of society.
      Values are statistics of the society and, in particular about the change of these over time.
      The selection process can take different forms, e.g., a public opinion poll or a representative's decision.

\end{enumerate*}

Clearly, other humans can be involved in the process of developing, implementing, and using \ac{adm} but are not our focus.
Below, we detail the decision-making process regarding the affected individuals.

\paragraph{Decisions about individuals.}

Decisions about individuals are derived according to the directed graphical model~\cite{koller2009probabilistic} in Figure~\ref{fig:framework}.
It consists of the following constituents:
\begin{enumerate*}[label=(\alph*), font=\itshape]
    \item \emph{Individuals.}
      Society is composed of individuals.
      We denote the state of an individual $i$ at time $t$ by random variable $S_t^i$.
      The state is to be understood to contain all information about an individual.
      We assume that individuals/their states are sampled from a distribution such that $S_t^i \sim p(S_t)$.
      
    \item \emph{Individual's observed and unobserved features.} 
      An individual's state $S_t^i$ yields a collection of observable features $X_t^i$ of that individual which are used as input to the algorithm as part of \ac{adm}.
      The features can for instance include demographic information or information about education and previous employment.
      There are also latent confounders $H_t^i$ which can be understood as all information about a person which is not available through the features $X_t^i$.
      For instance, confounders can contain information not permitted to be processed like race or gender.
      
    \item \emph{Algorithm's predictions.}
      The algorithm $\alg$ makes a prediction $P_t^i$ about the individual $i$ based on its features $X_t^i$, e.g., the probability that an unemployed person will find employment within 3 months.
    
    \item \emph{Algorithm's treatment recommendation.}
      Based on the prediction, the algorithm makes a treatment recommendation $\tilde{T}_t^i$, e.g., whether a person should receive further training.
    
    \item \emph{Actually applied treatment.}
      A \ac{pdm} reviews the algorithm's recommendation.
      Based on the algorithm's prediction $P_t^i$, its treatment recommendation $\tilde{T}_t^i$, and confounders $H_t^i$, the \ac{pdm} will make a final treatment recommendation $T_t^i$. 
      The \ac{pdm} can make use of confounders by intent, e.g., leverage information not provided to algorithm $\alg$, or unconsciously, e.g., because of biases regarding race or gender. 
      Sometimes we will refer to the mapping of a person's features to the applied treatment as \emph{treatment policy}.
    
    \item \emph{Outcome.} 
      The applied treatment $T_t^i$ together with the state $S_t^i$ of an individual will determine the outcome manifested as the state of the person at time $t+1$, i.e., $S_{t+1}^i$.

\end{enumerate*}
There can be additional confounders $H_t^\textnormal{SC}$ not related to an individual, e.g., changes in the law regarding care obligations.
For simplicity, we neglect these confounders.

\paragraph{Values.}

The \ac{sdm} selects an algorithm $\alg$ that best solves a problem at hand, e.g., predicting employment chances, and realizes its desired values. 
We assume that values correspond to statistics of the society at time $t$ and $t+1$, e.g., the change in conditional expectation of women being unemployed. 
Thus the \ac{sdm}'s aim is to select an algorithm $\alg^*$ that maximizes (or minimizes) an objective in combination with a weighted combination of statistics $\statistics_1, \ldots, \statistics_l$, i.e., $\alg^* \in \arg\max_{\alg \in \algs} F(\alg)$, where $F(\alg) = g(\alg) + \sum_{m=1}^l w_m \statistics_m(\alg)$, $g$ is the objective, and where the weights $w_m$ depend on the relative importance of social and ethical values.
More generally, the actual objective could be a complicated non-linear function of the respective statistics.

Clearly, ultimately the goal of the \ac{sdm} is likely to transform society, and the goals of this transformation might change.
If the goals don't change, values could also be expressed as desiderata on the stationary distribution of $S_t^i$.
This aspect has been studied in limited settings and regarding aspects orthogonal to our investigations, e.g.,~\cite{zhang2020fair}.
In our experiments, we limit ourselves to studying the impact of \ac{adm} in a single time step.

\vspace{-1mm}
\section{Stakeholders in the Face of Uncertainty}
\label{sec:experiments}

In order for the \ac{sdm} to pick the right algorithm, and for the \ac{pdm} to support the \ac{sdm} in realizing the desired values, each stakeholder needs complete knowledge of each other. %

\subsection{Control and Distribution of Power}
\label{sec:theory-control}

\paragraph{The \ac{sdm}'s perspective.}
While the \ac{sdm} should arguably have the power to decide upon the long-term objectives and pick the algorithm which best realizes them, it doesn't have this power in the face of (unconstrained) \acp{pdm}.
\begin{observation}
    In the face of unconstrained \acp{pdm}, the \ac{sdm} has no control over the realized objective/values.
\end{observation}
This observation is straightforward---without constraints, the \ac{pdm} can employ an arbitrary treatment policy. Nevertheless, it highlights that the \ac{sdm} should consider the \ac{pdm} when deciding upon the algorithm to use.

The above observation can be generalized to highlight that even for the case of constrained \acp{pdm}, there exist instances of our framework in which the \ac{sdm} makes a suboptimal choice when not accurately accounting for the \acp{pdm}.
\begin{observation}[]
   \label{thm:suboptimal}
    Assume an instantiation of the presented framework in which the \ac{pdm} is constrained to alter the algorithm's treatment recommendation in at most an $\epsilon$ fraction of all cases. 
    There exists an instantiation in which the \ac{sdm} does not select the optimal algorithm (from its perspective) if it does not account for the \ac{pdm}'s actions.
\end{observation}
Proof of the above statement is provided in Appendix~\ref{sec:appendix-proof}.
In practice, the exact \ac{pdm}'s behavior will typically not be known and, hence, the \ac{sdm} must make its choice facing uncertainty in that regard.
To this end, the \ac{sdm} could aim to make a robust choice regarding the algorithm to be used, e.g.:
\begin{align}
    \alg^* = \arg \max_{\alg \in \algs} \min_{\pi \in \Pi(\alg)} F(\pi), \label{eq:robust-algorithm}
\end{align}
where $\Pi(\alg)$ is the set of treatment policies the \acp{pdm} can realize when algorithm $\alg$ is used.
I.e., algorithm $\alg^*$ that maximizes $F(\cdot)$ assuming worst-case decisions of the \acp{pdm}.%

Importantly, the result from optimization problem~\eqref{eq:robust-algorithm} could also be used as an explanation of aspects of the decision-making problem to the \ac{sdm}, highlighting aspects regarding the robustness of the \ac{sdm}'s choice.

\paragraph{The PDM's perspective.}
Uncertainty about the SDM's values can limit the potential of the PDM. %
To see this, consider the (unrealistic) case that the \ac{pdm} had access to the true state of a person and could thus make perfect predictions, e.g., regarding their employment chances when being provided additional training.
However, in the face of uncertainty about the \ac{sdm}'s values, the \ac{pdm} might not be able to best leverage this predictive power in line with the \ac{sdm}'s intent:
if the \ac{sdm}'s goal is to maximize the chances that a person is employed, the \ac{pdm} could just impose a threshold on the employment chances to maximize this metric.
But the \ac{sdm} might also want to incentivize certain parts of the population to improve their qualifications and hence a simple threshold might not be sufficient.
Without appropriate information, the \ac{pdm} cannot act optimally.

In practice, additional issues might arise because of missing coordination among \acp{pdm} (if there are multiple): 
Different \acp{pdm} might be responsible for different demographic subgroups and to achieve a societal goal among subgroups, the \acp{pdm} must coordinate their actions, otherwise, their actions might be detrimental to the societal goals.

\subsection{Information Needs}
\label{sec:explanations}

To support the stakeholder's understanding of the decision-making process and its effects, different parts of the decision-making process must be explained. This is also illustrated by the dashed boxes in Figure~\ref{fig:framework}.
It has been shown that explanations can foster trust in \ac{adm}, help to identify biases, and support debugging of models.
Most common in the literature are explanations of  parts $E_1$ and $E_2$, directly studying aspects of the employed model/algorithm, i.e., explaining the predictions or treatment recommendations.
Many explanation methods have been particularly developed for this purpose including LIME, SHAP values, etc.---cf.\ \cite{molnar2022} for an overview.
While understanding explanations for $E_1$ and $E_2$ is important, they do not convey details important for the \ac{sdm} and \ac{pdm} to perform their tasks (cf.\ experiments). %
In particular, it is vital to provide explanations for $E_3$ and $E_4$, and, importantly, these explanations might need to cover different information for the \ac{sdm} and the \ac{pdm} as detailed below.
Additional information is presented in the appendix.

\paragraph{Explaining $\mathbf{E_3}$.}
Explaining $E_3$ includes the \ac{pdm}.
This is of crucial importance in our setting as the \ac{pdm} can alter the realized values and thereby counteract the strategic decisions made.
It is thus important to clearly explain the operation of the \ac{pdm} to the \ac{sdm} (or have the operation of the \ac{pdm} specified by the \ac{sdm} in a very narrow way).
Unfortunately, explaining the role of the \ac{pdm} is particularly challenging as in many cases the \ac{pdm} can consider confounding information, e.g., the appearance of a person that applies for support at the local employment center.
This confounding information can enable the \ac{pdm} to take \emph{better} actions as compared to the employed algorithm but can also be used in a way that counteracts the \ac{sdm}'s intent, e.g., because of prejudice which prevents changing biases in society like lower employment rates in certain subpopulations.
But also the \ac{pdm} must understand the \ac{sdm}'s values to decide optimally as, e.g., intended societal changes might not be in line with local optimal decisions.
      
\paragraph{Explaining $\mathbf{E_4}$.}
Explaining $E_4$ additionally includes the anticipated change in society on top of what is explained in $E_3$.
This is of particular importance for achieving the \ac{sdm}'s intended long-term changes.
Importantly, also the \ac{pdm} must understand those changes as their decisions, even if they were aligned with the \ac{sdm}'s long-term goals on short term (e.g., for a single time step), might have adversarial effects regarding long-term changes in society, e.g., because the dynamics form a negative feedback loop. %
Explanations $E_4$ require means to forecast changes in society, e.g., through models or simulations, and in many cases require carefully dealing with and conveying uncertainties inherent to these forecasts.
The forecasts would depend on the actions of the \ac{pdm} but can also depend on additional confounding factors, e.g., changes in politics beyond the concretely considered problem.

\vspace{-1mm}
\section{Experiments}

\subsection{Experimental Setup}

We support our findings by experimental results on a benchmark dataset, quantifying the impact of differing societal values and illustrating issues due to unmet information needs.
Additional information is available in the appendix.

\paragraph{Data and societal values.} 
We consider an artificial decision-making problem based on the well-known Boston housing dataset~\cite{harrison1978hedonic} to ensure easy reproducibility of our results.
To this end, we transform the Boston housing price prediction problem, which was originally a regression problem, into a classification problem by thresholding the regression target (``Median value of owner-occupied homes in \$1000's'') using the median as the threshold.
We consider the attributes ``proportion of blacks'' and ``property tax rate'' as sensitive attributes which should not influence a model's prediction (see details about how these are used in the prediction model below).
The dataset consists of 506 samples and has 13 features.

\paragraph{Prediction Models.}
We use logistic regression classifiers~\cite{bishop2006pattern} with fairness regularization for (approximately) achieving demographic parity~\cite{calders2009building} regarding different sensitive features.
Demographic parity requires that decisions are independent of the sensitive attributes which can be related to societal goals, e.g., making the provision of qualification measures for unemployed people fair regarding certain subpopulations.
We emphasize that the specific choice of machine learning model and quantification of societal values is not important and we use the considered model simply as a showcase---other model choices could clearly be sensible but are expected to lead to similar insights.
Concretely, given a labeled dataset $\mathcal{D} = \{(\x^{(i)}, y^{(i)})_{i=1}^n\}$, where $n$ is the number of data points, $\x^{(i)}$ and $y^{(i)} \in \{0,1\}$ the features and label of the $i$th sample, respectively, and statistics $\statistics_1, \ldots, \statistics_m$, we train a classification model with parameters $\theta$ to solve
\begin{align}
    &\max_\theta \sum_{i=1}^n \log p(y^{(i)} | \x^{(i)} ,\theta) - \sum_{l=1}^m w_l \rho_l(\theta), \label{eq:objective}
\end{align}
where
  $\rho_l = | \frac{1}{n} \sum_{i=1}^n (x^{(i)}_l - \bar{x}_l) d_\theta(\x^{(i)}) |$,
 $p(y^{(i)} | \x^{(i)}, \theta)$ is a generalized linear model with parameters $\theta$, $d_\theta(\x^{(i)})$ denotes the model's prediction for $\x^{(i)}$, $l$ denotes the $l$th sensitive attribute, $\bar{x}_l$ its mean, and $w_l > 0$.
The terms $\rho_l$ encourage the sensitive attributes to be equal on average to the population mean for samples classified as $1$, promoting fairness regarding the feature. 
The resulting optimization problem is convex and can be solved efficiently.

\subsection{Experimental Results}
\label{sec:experiments-results}

\paragraph{Mismatch of preferences.}

In our first experiments, we illustrate how much the \ac{pdm} can \emph{blur} the values the \ac{sdm} aims to realize.
In particular, we consider the case where the \ac{pdm} is represented by a classifier trained with weights $\tilde{\w}=[\tilde{w}_1, \ldots, \tilde{w}_l]$ which differ from the \ac{sdm}'s intended values $\w^*=[w^*_1, \ldots, w^*_l]$.
We are interested in understanding which values $\w$ would have led to classifiers with similar decisions as those from $\w^*$, i.e., for which the classifier's decisions differ only in some fixed percentage of cases.

\begin{figure}[!tb]
\begin{minipage}{0.49\textwidth}

  \centering
  
  \begin{subfigure}[c]{0.48\textwidth}
    \centering
    \includegraphics[width=3.8cm]{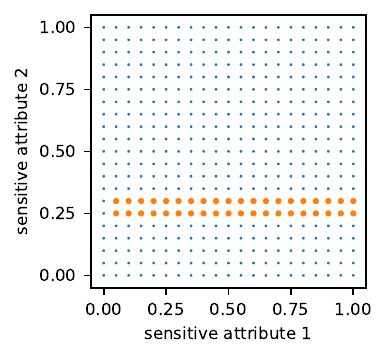}
    \vspace{-1mm}
    \subcaption{Appr.\ decision equivalence}
    \label{fig:possible-values-scatter}
  \end{subfigure}%
  \hfill
  \begin{subfigure}[c]{0.48\textwidth}
    \centering
    \includegraphics[width=3.8cm]{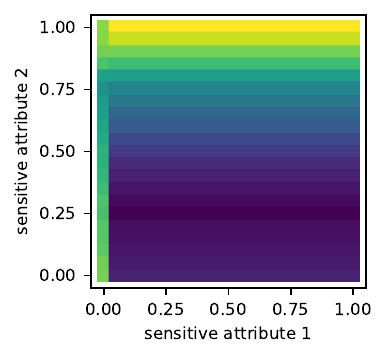}
    \vspace{-1mm}
    \subcaption{Deviation of decisions}
    \label{fig:possible-values-heatmap}
  \end{subfigure}
  
  \vspace{-1mm}
  \caption{Different possible weightings for societal values can lead to the same decisions if the \ac{pdm} can make decisions that deviate from the algorithm's recommendation.
  Sensitive attribute 1 is ``proportion of blacks by town'' and sensitive attribute 2 is ``full-value property-tax rate per \$10,000''.
  (\subref{fig:possible-values-scatter}) Different values realizing the same decisions in up to $1\%$ of the cases. We can understand this as the \ac{pdm} blurring the \ac{sdm}'s intent, or the \ac{sdm} giving up part of its power in favor of control of the system through the \ac{pdm}.
  (\subref{fig:possible-values-heatmap}) Sum of different decisions for deviations of the \ac{sdm}'s and \ac{pdm}'s values.
  }
  \label{fig:possible-values}

\end{minipage}%
\hfill
\begin{minipage}{0.49\textwidth}
  \centering
  \begin{subfigure}[c]{0.48\textwidth}
    \centering
    \includegraphics[width=3.8cm]{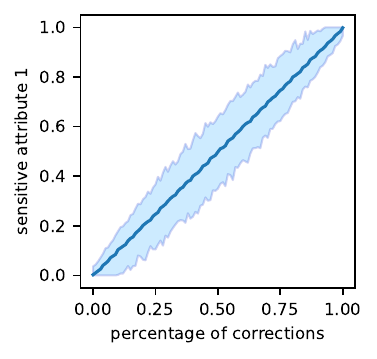}
    \vspace{-1mm}
    \subcaption{Change in the statistic for sensitive attribute 1}
  \end{subfigure}%
  \hfill
  \begin{subfigure}[c]{0.48\textwidth}
   \centering
   \includegraphics[width=3.8cm]{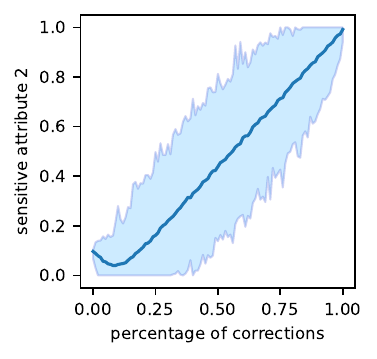}
   \vspace{-1mm}
   \subcaption{Change in the statistic for sensitive attribute 2}
  \end{subfigure}
  
  \vspace{-1mm}
  \caption{Change of realized societal values when correcting the recommended actions using the ground truth. A degradation of  values can be observed for an increasing number of corrections which can be attributed to the bias in the dataset.
  Sensitive attribute 1 is ``proportion of blacks by town'' and sensitive attribute 2 is ``full-value property-tax rate per \$10,000''.}
  \label{fig:degradation}
  \vspace{-3mm}
\end{minipage}
\end{figure}

We present such a result in Figure~\ref{fig:possible-values} for a threshold $\tau=1\%$ and \ac{sdm}'s weights $\w^* = [0.5, 0.25]$.
We observe that even for this low threshold a wide range of different other values would have led to similar decisions and the actually realized values are determined by the \ac{pdm}.
This indicates that, as expected and as intended, a certain level of control, and thus power, is \emph{transferred} from the \ac{sdm} to the \ac{pdm}.
Clearly, this can be acceptable if the realized values are roughly aligned with the \ac{sdm} and steer the society in the desired direction but can be problematic in other cases.

\paragraph{Degradation of social values when correcting decisions.}
To understand the impact of a \ac{pdm} correcting decisions in line with the ground truth, e.g., because of their expertise they make decisions in line with the outcomes of the past, we conducted an experiment in which we correct an increasing number of actions recommended by the model.
We use the same values and weights as before, i.e., $\w^* = [0.5, 0.25]$, perform corrections randomly, and average the impact on the statistics over 1000 runs.
Our results are shown in Figure~\ref{fig:degradation}.
We observe that, as expected, the corrections influence the statistics (the statistics/their change is normalized to the range $[0,1]$).
Importantly, in line with the experiments in the previous section, there is a large variance in the statistic for each value for some fixed percentage of corrections. 
This indicates that understanding the impact of a \ac{pdm} with domain expertise acting to the best of their knowledge can severely impact the realized societal values.

\paragraph{Local decisions.} %
In many real-world settings, for example in the case of the employment service algorithm, the \acp{pdm} would be caseworkers situated in local proximity to the individuals applying for support.
Thus we investigate whether different local decision strategies, in the sense of being applied to different subpopulations, impact the overall realization of societal values.
To this end, we consider a partition of the population according to societal characteristics and consider different strategies the \acp{pdm} could apply to these subpopulations.
In particular, we consider the following 3 stylized strategies:
\begin{enumerate*}[label=(\roman*), font=\itshape]
    \item \textsc{GlobalOpt}: Applying the optimal treatments based on the goals the \ac{sdm} wants to realize.

    \item \textsc{LocalCorrectGT}: The \acp{pdm} correct $p$\% of the decisions in the sense of selecting the `optimal' action (ground truth label). 

    \item \textsc{LocalOptSociety}: The objective which the \ac{sdm} wants to realize is optimized on the particular subpopulations the \acp{pdm} are facing.
    This corresponds to a setting in which the \acp{pdm} can be seen as experts for their domain and area, understand the \ac{sdm}'s intent, and apply them to the subpopulation they are interacting with.
\end{enumerate*}

To evaluate the strategies we artificially split the data into two subpopulations based on the sensitive attribute "proportion of blacks" (as threshold we used the value $50$).
Furthermore, we split data into training data ($80\%$) and test data ($20\%$).
We then applied the strategies to each of the splits (\textsc{LocalCorrectGT}, \textsc{LocalOptSociety}) or the whole data (\textsc{GlobalOPT}) and report the prediction accuracies as well as the values $\rho_l$ in Table~\ref{tab:vals}. 

\begin{table}[]
    \centering
    \scalebox{0.8}{
    \begin{tabular}{cccc}
        \toprule
        Strategy & sensitive attr.\ 1 & sensitive attr.\ 2 & accuracy \\\midrule
        \textsc{GlobalOPT} & $6.65 \pm 0.58$ & $3.40 \pm 0.24$ & $0.45$ \\
        \textsc{LocalCorrectGT} & $7.99 \pm 0.56$ & $3.56 \pm 0.28$ & $0.55$ \\
        \textsc{LocalOptSociety} & $5.94 \pm 0.53$ & $3.88 \pm 0.29$ & $0.48$ \\
        \bottomrule
    \end{tabular}}
    \caption{Effect of multiple \acp{pdm} on unseen data. If no global coordination among \acp{pdm} is ensured, local strategies can result in reduced realization of the desired values (larger values/penalizers in columns "sensitive attr.\ 1" and "sensitive attr.\ 2") or worse trade-offs among objectives (column "accuracy") and desired values. $\pm$ indicates standard errors.}
    \label{tab:vals}
    \vspace{-3mm}
\end{table}

The strategy \textsc{GlobalOpt} represents the best trade-off between goal-achievement and values that the \ac{sdm} wants to realize.
Interestingly, local strategies, i.e., \textsc{LocalCorrectGT} and \textsc{LocalOptSociety} will impact the desired trade-off in a non-trivial way if statistics depend on the global population or aspects of subpopulations which are shared among different \acp{pdm}.
This is clearly the case in most real-world settings that are characterized by our setting.
We can make our argument more formal by considering a set of $K$ \acp{pdm}, where the subpopulation \ac{pdm} $k$ ($1 \leq k \leq K$) is interacting with can be characterized by a distribution $p_k(\x, y)$.
The global population is then characterized by a distribution $p(\x,y)$ which is a mixture of the distributions $p_k(\x,y)$.
The statistics characterizing the values depend on $p(\x,y)$ in some way and if a \ac{pdm} optimizes societal values (or any other form of an objective) based on its local distribution $p_k(\x,y)$ there is, in general, no guarantee that the combination of the $K$ actions of the \ac{pdm} corresponds to the solution that we would be obtained when finding optimal treatments for $p(\x,y)$.
Most commonly considered fairness metrics will be affected by this issue as their constraints might depend on individuals from different subpopulations.
As a consequence, the \acp{pdm} can not achieve the overall goal using only \emph{local information} but require some form of \emph{global coordination}.

\vspace{-1mm}
\section{Implications, Discussion, and Ethics}
\label{sec:discussion}

In this section, we discuss implications of our results regarding distribution of power, information needs, and limitations.

\paragraph{Distribution of power.}
While the power to decide on the (long-term) goals should lie solely with the \ac{sdm}, this is in general not true if \acp{pdm} are involved.
This results in a tension between what the \ac{sdm} wants to achieve and what the \ac{pdm} is supposed to do, as also emphasized by our experiments which illustrate the possible change in values because of the \ac{pdm}.
On the one hand, this tension requires also thinking of the \ac{pdm} as a strategic and political decision-maker to some extent, defining a clear scope and clear limits for \ac{pdm}'s actions, and carefully reviewing the reasons if a \ac{pdm} takes actions which deviate from the algorithm's recommendations frequently.
On the other hand, the \acp{sdm} are required to be transparent decision-makers that communicate their values and goals and support the \acp{pdm} in fulfilling their roles.%

\paragraph{The stakeholders and their struggles.}
Both the \ac{sdm} and the \ac{pdm} are part of and interact with a socio-technical system that has values embedded. 
Aligning the \ac{sdm} and the \ac{pdm} requires transparency about those values.
But \cite{boddington_2017_towards} raises the question of whether we can fully articulate our most fundamental values.
Expecting the \ac{pdm} to make decisions, which are better than the algorithm’s decision must happen with respect to the \ac{sdm} and should intend to compensate for modeling errors or account for latent confounding factors.
Still, the \ac{pdm}'s decisions might be personally colored and not aligned with strategic goals or public agenda. 
We, therefore, believe the potential influence of the \ac{pdm}'s (personal) values is under-discussed.

While the \ac{sdm} might not have as much power as they should, the \ac{pdm} might have too much in some sense while struggling to utilize it.
The \ac{pdm} might also be as powerful as implied in policy frameworks, like the AI Act~\cite{europeancommission_2021_artifical}. 
Being a \ac{hil} is a blurry and not clearly defined position with different and changing expectations. It is a position that can shape the \ac{adm}'s behavior but might not fully be able to account for all effects and the value systems it affects.
Having more power to adapt the \ac{adm} might avoid bias or discrimination, but also causes issues regarding responsibility and its assignment~\cite{coeckelbergh_2020_ai}.

In cases in which the values that the \ac{adm} should realize are decided upon in democratic processes, this brings up a contrast between technocracy as expert rule and democracy as a people rule. 
We would need to rethink how and why algorithms are commissioned if we want to question the power of the \ac{sdm}.
To clarify, here, we treat the \ac{sdm} as someone who is deciding over the direction of the algorithm, which can imply all sorts of algorithms that are democratically influential. %
This does not mean that their value system is translated or implemented one to one, but it means that this is a top-down technocratic model to assign algorithms.

Finally, considering the role of the \ac{pdm} we also have to take potential detachment effects of \ac{adm} into account which have been reported in~\cite{bader2019algorithmic}. %
People can become distanced when they feel removed from their basis of decision-making \cite{bader2019algorithmic}.
Overcoming this would require a more in-depth engagement with the person applying for support, their data, and the subsequent translation to categories used for analysis and collective information processing \cite{scott_algorithmic_2022,DBLP:conf/nordichi/MollerSH20}. 

\paragraph{Explanations for the stakeholders.}
For the \ac{sdm} and the \ac{pdm} to realize their goals and fill their roles their differing information needs (= explanation needs) must be satisfied. 
In particular, the \ac{sdm} must be able to understand the \ac{adm}'s impact in combination with the \acp{pdm} while the \acp{pdm} need to understand the values the \ac{sdm} wants to realize as well as the dynamics of the decision process to take corresponding actions.
Such difference in explanation needs is in line with recent work, e.g., \cite{DBLP:conf/ACMdis/DhanorkarWQXP021}, which argues to think of explanations as iterative, interactive, and emergent.

A concrete approach for explanations for the \ac{sdm} could be informed by the results presented in our experiments, which illustrate how values are blurred because of the actions the \acp{pdm} can take.
Furthermore, we could produce similar explanations for interacting with \acp{pdm} who might have a different value system (which might be known partially because of experience or appropriate studies) and thus understand the risks imposed by the \ac{pdm} with respect to the realization of the values.
It would also be possible to perform a sort of value inference based on the actual decisions of a \ac{pdm} which could then be checked against the intended values of the \acp{sdm} and used to better instruct particular \acp{pdm}.

Regarding concrete suggestions for the development of explanations, for the \ac{pdm}, we have observed the need for \acp{pdm} to understand the societal values they should realize requiring appropriate communication.
Furthermore, we have seen the need for the \acp{pdm} to coordinate globally with their peers---this need could be met by actively, during the usage of the \ac{adm}, providing statistics of the currently made global decisions and suggestions of how to include this information in the local decision-making.

\vspace{-1mm}
\section{Conclusions}
\label{sec:conclusions}

We considered the role of humans in \ac{adm} for socially relevant problems, focusing on the \acfp{sdm} making long-term strategic decisions and the \acfp{pdm} making decisions about the treatment of individuals.  
After conceptualizing the considered setting, we illustrated tensions arising from diverse expectations, values, and constraints by and on the humans involved and exemplified them in experiments on a machine learning benchmark dataset.
We found that while the \ac{pdm} is typically assumed to be a corrective, they can, in theory, counteract the realization of the societal values, not least because of a misalignment of these values and unmet information needs.
This can lead to a transfer of power from the \ac{sdm} to the \ac{pdm} and should be accounted for in explanations of the \ac{adm}.
Suitable explanations should emphasize the \ac{pdm}'s implicit role as a strategic decision-maker.

\subsubsection{Acknowledgments.}
This work has been funded by the Vienna Science and Technology
Fund (WWTF) [10.47379/ICT20058] as well as [10.47379/ICT20065].

\bibliographystyle{named}
\bibliography{references}

\clearpage
\appendix

This appendix contains:
\begin{itemize}
    \item Further information on the conducted experiments in Section~\ref{sec:experimental-details}.
    
    \item Further notes on the design implications for explanations resulting from our considerations~\ref{sec:design-implications}.
    
    \item Further information on implementing the optimization objective in Equation~\eqref{eq:objective} in real-world situations in Section~\ref{sec:optimization-objective}.

    \item Additional related works in Section~\ref{sec:related-extended}.

    \item The proof of Observation~\ref{thm:suboptimal} in Section~\ref{sec:appendix-proof}.
\end{itemize}

\section{Additional Experimental Details and Results}
\label{sec:experimental-details}

\textbf{Objective.} The objective in Equation~\eqref{eq:objective} serves as an example objective the \acp{sdm} might aim top optimize. 
Note, however, that such an objective might not be available explicitly (cf.\ Section~\ref{sec:optimization-objective}) and within this paper it simply serves to illustrate the effects the misalignment between \acp{sdm} and \acp{pdm} can have and illustrate examples of how such a misalignment can be explained through visualizations.
For the sake of simplicity we chose demographic parity as the means to quantify values of the \acp{sdm} but in practice values could of course be more general and could be measured by other fairness metrics or quantified by other measures the \acp{sdm} consider relevant.

In Equation~\eqref{eq:objective}, the terms
$$\rho_l = | \frac{1}{n} \sum_{i=1}^n (x^{(i)}_l - \bar{x}_l) d_\theta(\x^{(i)}) |$$ quantify the average deviation of the $l$th sensitive attribute from its mean weighted by the model's predictions.
Thus these terms encourage that for $1$-decisions, the sensitive attributes are close to their mean, i.e., individuals are not preferred because of their sensitive attributes.

\textbf{Choice of dataset.} 
We used the well-known publicly available housing dataset (based on the US census) to ensure easy reproducibility of our results.
Furthermore, most public employment datasets do not contain information about individuals.
One exception is the Kaggle dataset \emph{Students’ Employability - Philippines}\footnote{\url{https://www.kaggle.com/datasets/anashamoutni/students-employability-dataset}} but we could not find detailed information about how the data was collected and the precise meaning of the provided binary target label (and how precisely it was derived).
Thus, we decided to use the more standard dataset.

\textbf{Implementation.} 
Our classifier implementation is building on \texttt{scikit-lego}\footnote{\url{https://pypi.org/project/scikit-lego/}}.
For reference, easy reproducibility and extension of our results, our source code is provided in the supplemental material.

\section{Design Implications for Designing Explanations}
\label{sec:design-implications}

\textbf{On \acp{sdm} and \acp{pdm}.}
The \acp{sdm} and \acp{pdm} can be the same person, but often this will not be the case. For instance, an organization’s managers could decide to introduce an AI system to make some decision-making processes more effective, e.g., for deciding on loan applications or college admissions.
These managers then will not use these systems themselves but rather the staff of their respective organizations.
This is precisely the reason why there can be a mismatch in intent. 

\textbf{Explanation design implications.}
A key observation of our paper is that for the considered setting both the \acp{sdm} and the \acp{pdm} will often require tailored explanations to perform their tasks well and achieve their goals.
For instance, while $E_1$ and $E_2$ in Section~\ref{sec:explanations} can in principle be explained using classical explainability techniques (cf.\ \cite{molnar2022}), these explanations do not cover downstream effects including the interplay and alignment of the \acp{sdm} and the \acp{pdm} regarding the actual decisions that would be made.
But these aspects can be crucial for the \acp{sdm} and \acp{pdm} to perform their tasks well and achieve their goals.
This has immediate consequences on which explanations should be used and how they should be designed. 
On the one hand, if explanations for $E_1$ and $E_2$ are presented to the \acp{sdm}, they could for instance either
\begin{enumerate*}[label=(\alph*)]
    \item include a hint highlighting that the interpretation of $E_1$ and $E_2$ could be blurred if the \acp{pdm} were included or
    \item try to account for the \acp{pdm}, e.g., by assuming some worst-case behavior of the \acp{pdm}\footnote{for instance resulting from imposing constraints on them, e.g., regarding the number of decisions they can make that do not agree with the algorithm's recommendation} and artificially modifying $P_t^i$ or $\tilde{T_t^i}$ to be consistent with the decisions the \acp{pdm} would make.
\end{enumerate*}
On the other hand, if explanations for $E_1$ and $E_2$ are presented to the \acp{pdm}, they could for instance also include a hint as above or be augmented with exemplary cases that illustrate when the \acp{sdm} would expect the \acp{pdm} to make decisions that deviate from those that would result from $P_t^i$ or $\tilde{T_t^i}$ so that the \acp{pdm} can better understand the intent (values) of the \acp{sdm}.

Similar design implications can be derived for $E_3$ and $E_4$ with the addition of also including a dynamics model, e.g., by building on insights from economics and the social sciences to illustrate the expected future impact of decisions. But there are also many more options, e.g., for $E_4$ the forecasted changes in employment for various age groups for different possible behaviors of the \acp{pdm} could be visualized so that the \acp{sdm} can make a more informed selection of the algorithm $\alg$ they want to deploy.

\section{On Implementing the Optimization Objective in Real-world Situations}
\label{sec:optimization-objective}

The optimization objective in Equation~\eqref{eq:objective} mainly serves as a motivational example of the goals the \acp{sdm} might pursue.
In many real-world situations, the \acp{sdm} might not make such an objective explicit, not least because articulating values is difficult.
Nevertheless, and also for the sake of transparency and to support explainability and accountability, \acp{sdm} could aim to at least approximately spell out such an objective.
Consider the employability prediction problem.
One could adopt an iterative process as values can be difficult to articulate.
The \acp{sdm} could be interviewed regarding the values they aim to realize, e.g., "as many people in employment as possible", "equal chances for different groups", $\ldots$, and their relative importance.
From this information, a proposal for the objective can be developed, and an algorithm that maximizes it (using another model for the dynamics, e.g., coming from economics).
One could then refine this model with the \acp{sdm} until it sufficiently accurately reflects their values. 

But even if values cannot be quantified, AI-supported \ac{adm} will be realized and used.
If in such a case a large number of decisions is modified by the \acp{pdm}, interviews as well as further analysis could be conducted to understand the intent of the \acp{pdm}’ actions and verify these against the goals with which \ac{adm} was introduced.

\section{Extended Related Work}
\label{sec:related-extended}

\subsection{Humans-in-the-loop}

A \emph{human-in-the-loop} setting is commonly defined as an automated process that requires human interaction, meaning that human knowledge and experience are integrated into, for instance, an ML model~\cite{DBLP:journals/fgcs/WuXSZM022}.
While there is no clear-cut definition, some authors also differentiate the \emph{human-on-the-loop} process in which the human plays a less central role; referring to tasks that are executed completely and independently by machines, but where a person acts in a monitoring or supervisory role, with the ability to interfere if the process should fail~\cite{fischer2021loop}.
Another term, often used interchangeably, is \emph{human oversight} often discussed in the context of trust in \ac{adm} and AI and which is increasingly required by policy-makers and regulators~\cite{koulu2020human,high_level_expert_group_on_ai_eu,europeancommission_2021_artifical}. 
As mentioned, we investigate two specific types of \acp{hil} in this work, which both influence \ac{adm}'s outcomes:
\begin{enumerate*}[label=(\roman*), font=\itshape]
  \item a person who makes a final determination of a decision informed by the system’s predictions~\cite{bell2022itsjust}---the \emph{\acf{pdm}} or \textit{practical \ac{hil}}; and

  \item a person who has strategic oversight over the development and deployment of an \ac{adm}, such as politicians, regulators, or other actors in society who determine longer-term goals in the context of public services---the \emph{\acf{sdm}} or \textit{strategic \ac{hil}}.
\end{enumerate*}

The \ac{hil}, generally speaking, is commonly viewed as a control system that guarantees accountability but also prevents undesirable consequences~\cite{dodge_explaining_2019,starke2022fairness_perceptions,high_level_expert_group_on_ai_eu,rahwan2018society}. 
Both from a legal and a moral fairness perspective the role of the \ac{hil} is to mitigate risk.
Algorithmic accountability has been established as a term to describe good practice for \ac{adm} use in public services~\cite{brown_toward_2019}, often attributed to the existence of a \ac{hil}.
However, the traditional \ac{hil}, deeply embedded in the technicalities of the ADM process does not fit this role necessarily.
Usually, the focus is on increasing the performance of the algorithm, resolving ambiguities, or even simply labeling input data~\cite{DBLP:journals/fgcs/WuXSZM022}.
As said, in this work, we investigate two different \acp{hil} which influence the \ac{adm} process at different points, and which are of particular importance in \ac{adm} with societal impact.
They are illustrated in Figure~\ref{fig:framework-human-readble} as the \acf{pdm}, who oversees individual decisions (such as a caseworker), and the \acf{sdm} who is often a decider of values and long-term decisions.

Most research on \ac{adm} does not specify the precise role of the \ac{hil} nor of the \ac{pdm} or \ac{sdm} in detail, nor does it study it in abstract form, e.g., whether a \ac{hil} has the power to influence the \ac{adm} outcome and how this would affect the wider environment. 
Questions regarding the distribution of power are raised for example by ~\cite{kasy2021fairness}, who discuss who gets to select the objective function of an algorithm.
Describing the role of these \acp{hil} requires thinking about the work practices where a model’s predictions are meant to be used and the environment in which the \ac{hil} is expected to act~\cite{DBLP:conf/ACMdis/DhanorkarWQXP021}.
This is likely particularly relevant for the \ac{pdm} who is in some cases expected to use the \ac{adm} outcome simply as a second opinion~\cite{allhutter_bericht_ams-algorithmus_2020}.
We argue that the \ac{pdm} is put in a potentially unresolvable position of tension between individual decisions and strategic value-based desired outcomes.
The expectations of the \ac{pdm} are not aligned with their practical power or expected ability to understand strategy.
We challenge the assumption that the \ac{pdm} has safeguarding capabilities that are so readily assigned to this role. 

In light of the growing complexity of models, other authors have identified explanations of a system's output as necessary, next to being able to evaluate it against a set of norms, constraints, and standards~\cite{langer_what_2021,speith_review_2022}. But also the need to understand larger parts of the decision-making process has been recognized~\cite{kulesza2013too,madras2018predict}.
This is especially relevant as the \acp{hil} needs to understand algorithmic decisions to enact their role of oversight, thus explanations need to be available and accessible~\cite{hind2019ted}.
However, while recent work highlights the need to communicate \ac{adm} in different ways there are limitations in how we can explain and assess such values~\cite{wallach_2009_moral}.

Importantly, the influence of explanations on fairness perceptions and the resulting impact on human-AI decision making has been considered in related work~\cite{schoeffer2022explanations}.
While this work also considers how explanations can influence a \ac{hil}'s actions, it focuses on fairness perceptions and not on the more general alignment of arbitrary values we focus on.
Furthermore there is no conceptual distinction regarding the explanations of different parts of the AI assisted decision-making and thus questions of alignment are not discussed in the granularity as in our work. 
Further, our discussion considers philosophical aspects and ethical concerns and issues arising from the general problem setup and the challenge of alignment in \ac{adm} with a \ac{hil} in particular.

\subsection{Information needs of humans-in-the-loop}

Different \acp{hil} need different types of knowledge about the \ac{adm} process depending on their role~\cite{langer_what_2021}. 
To date, most \ac{xai} literature focuses on the technicalities of automated processes~\cite{speith_review_2022,abdul2018trends}. 
While recent work and guidelines demonstrate increasing awareness of the potential pitfalls and dangers of \ac{adm}~\cite{brown_toward_2019,definelicht2020}, many proposed solutions are designed with data scientists or developers in mind, closely linking explanations with the model process,~\cite{DBLP:conf/chi/LiaoGM20}. 
This common, algorithmic-centric view of information needs and explanations focuses on information about input data and potential biases, performance, feature weights, and other aspects of \ac{adm}~\cite{DBLP:conf/chi/LiaoGM20}.
While relevant, a need for other diverse information types has been recognized. Within the technical realm this includes considering `what' is explained; for instance, differentiating between \emph{local} (a predictive output is explained) and \emph{global}
(concerning the broader overall reasoning of the model) explanations
\cite{DBLP:conf/ACMdis/DhanorkarWQXP021}.
What is more, this also includes considering `who' explanations are tailored to, as different stakeholders in \ac{adm} have been identified, such as developers, users, and affected persons~\cite{langer_what_2021}.
This work takes a step further in illustrating the cross-disciplinary information needs of the \acp{hil}, widening the scope of required information needs, skills, and explanations; as well as an in-depth discussion of the role and power of two \acp{hil} directly involved in the \ac{adm} process, the \ac{pdm} and \ac{sdm}.

In light of the growing complexity of models, other authors have identified explanations of the output of a system as necessary, next to being able to evaluate it against a set of norms, constraints, and standards~\cite{langer_what_2021,speith_review_2022}. But also the need to understand larger parts of the decision-making process has been recognized~\cite{kulesza2013too}.
This is especially relevant as the \acp{hil} needs to understand algorithmic decisions in order to enact their role of oversight, thus explanations for the decision need to be available and accessible~\cite{hind2019ted}.
However, while, as already mentioned, recent work highlights the need to communicate \ac{adm} in different ways, including not just technical processes, but also acknowledging desired values, there are limitations in how we can explain and assess such values~\cite{wallach_2009_moral}.~\cite{winikoff_2018_towards} for instance discusses the complexities of `value-based reasoning' from algorithms and other autonomous systems, while others question in what sense a technological system `uses' or `represents' values at all~\cite{boddington_2017_towards}. Also, studying the empirical processes where value is assigned to algorithms is challenging~\cite{petersen2020role,DBLP:conf/nordichi/MollerSH20}. In the context of public services, \cite{DBLP:conf/nordichi/MollerSH20} illustrate the complexities of discussing values in \ac{adm} at an organizational level as well as an individual level with caseworkers, a differentiation which this work picks up and expands on.

Calls for explanations targeting different audiences have been made, not least to support public understanding (including affected persons) of \ac{adm}~\cite{woodruff_qualitative_2018}. 
It has also been recognized that the information needs of external stakeholders likely require different levels of detail than those directly involved in the process and that explanation needs are highly contextual and situated~\cite{DBLP:conf/ACMdis/DhanorkarWQXP021}.
A recognition of the dynamic nature of AI models and their embeddedness in sociotechnical systems has resulted in the realization that explanations might not be static information objects, at best tailored to different audiences, but facilitators of ongoing sensemaking and collaborative learning by different actors in the \ac{adm} process~\cite{DBLP:conf/ACMdis/DhanorkarWQXP021,sokol2020one}. 
In this paper, we illustrate the need for more complex explanations that actively aim to support resolving tensions between practical and strategic value-based decisions, considering the various constraints under which \acp{hil} operate.

Issues regarding \acp{pdm} being unaligned with policy goals have also been studied without relation to \ac{adm}~\cite{prottas1978power,lipsky2010street,pratt2009brief,sossin1994redistributing,kelly1994theories}.
Several works have also put this into the context of \ac{adm}~\cite{bovens2002street,alkhatib2019street,binns2022human}.
Most of these works, while addressing a similar topic as our paper and raising related issues, touch the topic at a higher level of technical detail but we argue that a rigorous treatment also on a technical systems level is necessary to deal with this topic. 
There is also fieldwork studying the interaction of \ac{adm} and \acp{hil}~\cite{cheng2022child} and related work taking a more holistic picture of the whole decision-making process touching upon the topics of \ac{xai} and issues regarding fairness~\cite{madras2018predict,bansal2021does}.

\subsection{Ethical responsibility in public service \ac{adm}}

Explaining decisions is not only a part of what humans naturally do when they communicate, but it is also a moral requirement~\cite{coeckelbergh_2020_ai}, arguable even more so for \ac{adm} with societal consequences. 
Decision-support in public service is seen as a domain in its own right~\cite{DBLP:conf/nordichi/MollerSH20,veale_2018_fairness}. 
One characteristic as pointed out by~\cite{DBLP:conf/nordichi/MollerSH20} is how challenging it can be to determine the success of interventions due to the long-term and procedural nature of the public service domain. 
Metrics that correspond to values are contested due to the complexity of societal processes, for instance concerning resource allocation in employment support.
Complex algorithms influence not only expert contexts like medicine~\cite{chung_2021_covidnet} but also increasingly democratic spheres~\cite{sudmann_2019_the}; another incentive to rethink the scope of information that \ac{xai} is addressing. 
In this light, understanding the power of the \ac{hil} is especially important as their role is concerning decisions that have a wider democratic scope.

Values are an inherent part of any automated decision-making process. For instance, Wallach \& Allen~\cite{wallach_2009_moral} point out the need to acknowledge that we program values into machines and that this makes them ethically consequential even if not yet fully agents. 
Since then, Vallor~\cite{vallor_2016_technology} and Coeckelbergh~\cite{coeckelbergh_2020_ai} have developed various ethical concepts to understand the development of virtues within technology or AI. 
\cite{spiekermann_value-lists_2021} advocate value-based engineering as a process-driven, holistic approach to system engineering. 
\ac{adm} in public services is explicitly value-driven~\cite{brown_toward_2019,veale_2018_fairness}. 
Following~\cite{spiekermann_value-lists_2021}'s approach we argue that if the \ac{hil}'s are considered an integral part of the \ac{adm} process in the public domain it is insufficient to provide them with `lists of values', but that we need to rethink the role, knowledge, and power of each \ac{hil} to ensure the consistency of value principles in the decision-making process. 
\cite{coeckelbergh_2020_ai} points out that the issue with responsibility is its assignment, a point which is emphasized by the challenges faced by the \ac{sdm} and the \ac{pdm}. %
Despite efforts to the contrary, several reasons have been identified why \ac{adm} systems currently rarely comply with standards that aim to ensure such value-driven design, such as for instance the standard for `trustworthy artificial intelligence'~\cite{high_level_expert_group_on_ai_eu,definelicht2020,bell2022itsjust,dodge_explaining_2019}. %
Much discussed values in this context have been fairness and interpretability of algorithms in \ac{adm} settings. 
This has led to auditing approaches of algorithms~\cite{feldman2015certifying,zafar2017fairness,bandy2021problematic} and the development of `fair’ by design model construction approaches~\cite{brown_toward_2019,agarwal2018reductions,kilbertus2017avoiding}. 
In this work, we discuss the importance of understanding and communicating how chosen values are operationalized in an \ac{adm} use case and the complex decision design space of public services~\cite{yu2020keeping}.

Describing the roles of the different \acp{hil} requires thinking about the work practices where a model’s predictions are meant to be used and the environment in which the \ac{hil} is expected to act~\cite{DBLP:conf/ACMdis/DhanorkarWQXP021}.

\section{Proof of Observation~\ref{thm:suboptimal}}
\label{sec:appendix-proof}

\begin{proof}
    We prove the statement by constructing an instance of our framework in which the \ac{sdm} selects a suboptimal algorithm regarding the \ac{sdm}'s objective if the \ac{sdm} does not take the \ac{pdm}'s actions into account.
    To this end, assume that $\algs=\{\alg_1, \alg_2\}$, i.e., the \ac{sdm} can choose between two different algorithms.
    Furthermore, assume that $F(\alg)=g(\alg)=\text{accuracy}(\alg)$, i.e., there are no values and the \ac{sdm}'s objective is to maximize $\text{accuracy}(\alg)$, e.g., the accuracy of $\alg$ for predicting whether a person will not find a job without additional training (the \ac{sdm}'s goal could be to provide training to all those people).
    Wlog let $F(\alg_2) < F(\alg_1) < F(\alg_2) + \delta$ for some $0 < \delta < \epsilon$ (this is always possible if the achievable accuracy for some machine learning problem is at least $\epsilon$ and full knowledge about the data distribution is available).
    That is, the \ac{sdm} prefers $\alg_1$ over $\alg_2$ as it achieves better performance wrt to its goals.
    But if the \ac{pdm} can alter $\alg_2$'s decisions on cases on which $\alg_2$ makes mistakes while it cannot correct mistakes for $\alg_1$, then the \ac{pdm} can together with $\alg_2$ achieve a performance of $F(\alg_2) + \epsilon > F(\alg_2) + \delta > F(\alg_1)$ and hence the choice of the \ac{sdm} of selecting $\alg_1$ would be suboptimal in face of employing the \ac{pdm}.
\end{proof}

\end{document}